\documentclass{article}
\usepackage[T1]{fontenc}
\usepackage{babel}
\usepackage{graphics}

\makeatletter

\providecommand{\LyX}{L\kern-.1667em\lower.25em\hbox{Y}\kern-.125emX\@}
\newcommand{\noun}[1]{\textsc{#1}}

 \usepackage{verbatim}

\usepackage{graphics}

\usepackage{amstex}
\usepackage{colacl}
\usepackage{rotating}
\usepackage{multicol}

\begin{document}

\title{Dynamic Nonlocal Language Modeling via \\
 Hierarchical Topic-Based Adaptation}

\author{Radu Florian \and David Yarowsky \\
Computer Science Department and Center for Language and Speech Processing,
\\
Johns Hopkins University\\
Baltimore, Maryland 21218\\
\{rflorian,yarowsky\}cs.jhu.edu}

\maketitle
\begin{abstract}
This paper presents a novel method of generating and applying hierarchical,
dynamic topic-based language models. It proposes and evaluates new
cluster generation, hierarchical smoothing and adaptive topic-probability
estimation techniques. These combined models help capture long-distance
lexical dependencies. Experiments on the Broadcast News corpus show
significant improvement in perplexity (10.5\% overall and 33.5\% on
target vocabulary).
\end{abstract}

\section{Introduction}

Statistical language models are core components of speech recognizers,
optical character recognizers and even some machine translation systems
\newcite{brown90}. The most common language modeling paradigm used
today is based on \emph{n-grams}, local word sequences. These models
make a Markovian assumption on word dependencies; usually that word
predictions depend on at most \emph{m} previous words. Therefore they
offer the following approximation for the computation of a word sequence
probability: {\small \( P\left( w_{1}^{N}\right) =\prod _{i=1}^{N}P\left( w_{i}|w_{1}^{i-1}\right) \approx \prod _{i=1}^{N}P\left( w_{i}|w_{i-m+1}^{i-1}\right)  \)}
where \( w_{i}^{j} \) denotes the sequence \( w_{i}\ldots w_{j} \)
; a common size for \emph{\( m \)} is 3 (trigram language models).

Even if n-grams were proved to be very powerful and robust in various
tasks involving language models, they have a certain handicap: because
of the Markov assumption, the dependency is limited to very short
local context. Cache language models (\newcite{kuhn92},\newcite{rosenfeld94})
try to overcome this limitation by boosting the probability of the
words already seen in the history; trigger models (\newcite{lau93}),
even more general, try to capture the interrelationships between words.
Models based on syntactic structure (\newcite{chelba98}, \newcite{wright93})
effectively estimate intra-sentence syntactic word dependencies.

The approach we present here is based on the observation that certain
words tend to have different probability distributions in different
topics. We propose to compute the conditional language model probability
as a dynamic mixture model of \( K \) topic-specific language models:
\vspace{-0.3cm}\begin{equation}
\label{12}
\begin{array}{r}
P\left( w_{i}|w_{1}^{i-1}\right) =\sum\limits _{t=1}^{K}P\left( t|w_{1}^{i-1}\right) \cdot P\left( w_{i}|t,w_{1}^{i-1}\right) \\
\approx \sum\limits _{t=1}^{K}P\left( t|w_{1}^{i-1}\right) \cdot P_{t}\left( w_{i}|w_{i-m+1}^{i-1}\right) 
\end{array}
\end{equation}

\begin{figure}[tb]
{\centering \resizebox*{0.5\textwidth}{0.25\textheight}{\includegraphics{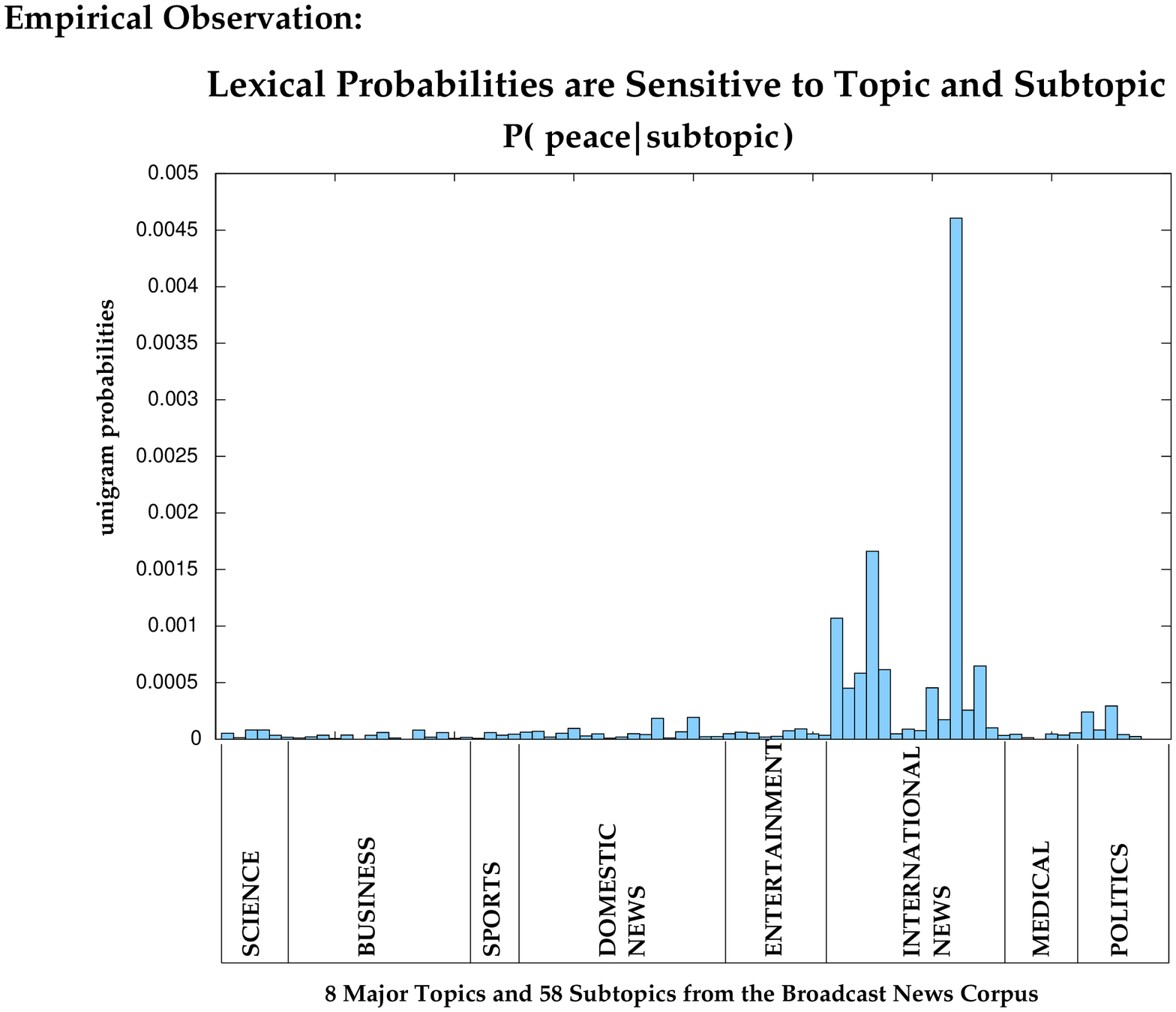}} \par}

\caption{Conditional probability of the word \emph{peace} given manually assigned
Broadcast News topics\label{peace}}
\end{figure}
\vspace{-0.3cm}The motivation for developing topic-sensitive language
models is twofold. First, empirically speaking, many \( n \)-gram
probabilities vary substantially when conditioned on topic (such as
in the case of content words following several function words). A
more important benefit, however, is that even when a given bigram
or trigram probability is not topic sensitive, as in the case of sparse
\( n \)-gram statistics, the topic-sensitive unigram or bigram probabilities
may constitute a more informative backoff estimate than the single
global unigram or bigram estimates. Discussion of these important
smoothing issues is given in Section \ref{TheLanguageModel}.

Finally, we observe that lexical probability distributions vary not
only with topic but with subtopic too, in a hierarchical manner. For
example, consider the variation of the probability of the word \emph{peace}
given major news topic distinctions (e.g. \emph{\noun{business}}
and \emph{\noun{international}} news) as illustrated in Figure \ref{peace}.
There is substantial subtopic probability variation for \emph{peace}
within \emph{\noun{international}} news (the word usage is 50-times
more likely in \noun{international:middle-east} than \noun{international:japan}).
We propose methods of hierarchical smoothing of \( P(w_{i}|\textrm{topic}_{t}) \)
in a topic-tree to capture this subtopic variation robustly.

\subsection{Related Work}

Recently, the speech community has begun to address the issue of topic
in language modeling. \newcite{lowe95} utilized the hand-assigned
topic labels for the Switchboard speech corpus to develop topic-specific
language models for each of the 42 switchboard topics, and used a
single topic-dependent language model to rescore the lists of N-best
hypotheses. Error-rate improvement over the baseline language model
of 0.44\% was reported.

\newcite{iyer94} used bottom-up clustering techniques on discourse
contexts, performing sentence-level model interpolation with weights
updated dynamically through an EM-like procedure. Evaluation on the
Wall Street Journal (WSJ0) corpus showed a 4\% perplexity reduction
and 7\% word error rate reduction. In \newcite{iyer96}, the model
was improved by model probability reestimation and interpolation with
a cache model, resulting in better dynamic adaptation and an overall
22\%/3\% perplexity/error rate reduction due to both components.

\newcite{seymore97} reported significant improvements when using a
topic detector to build specialized language models on the Broadcast
News (BN) corpus. They used TF-IDF and Naive Bayes classifiers to
detect the most similar topics to a given article and then built a
specialized language model to rescore the N-best lists corresponding
to the article (yielding an overall 15\% perplexity reduction using
document-specific parameter re-estimation, and no significant word
error rate reduction). \newcite{seymore98} split the vocabulary into
3 sets: general words, on-topic words and off-topic words, and then
use a non-linear interpolation to compute the language model. This
yielded an 8\% perplexity reduction and 1\% relative word error rate
reduction.

In collaborative work, \newcite{Mangu97} investigated the benefits
of using existing an Broadcast News topic hierarchy extracted from
topic labels as a basis for language model computation. Manual tree
construction and hierarchical interpolation yielded a 16\% perplexity
reduction over a baseline unigram model. In a concurrent collaborative
effort, \newcite{khudanpur99} implemented clustering and topic-detection
techniques similar on those presented here and computed a maximum
entropy topic sensitive language model for the Switchboard corpus,
yielding 8\% perplexity reduction and 1.8\% word error rate reduction
relative to a baseline maximum entropy trigram model.

\section{The Data}

The data used in this research is the Broadcast News (BN94) corpus,
consisting of radio and TV news transcripts form the year 1994. From
the total of 30226 documents, 20226 were used for training and the
other 10000 were used as test and held-out data. The vocabulary size
is approximately 120k words.

\section{Optimizing Document Clustering for Language Modeling\label{optimizesec}}

For the purpose of language modeling, the topic labels assigned to
a document or segment of a document can be obtained either manually
(by topic-tagging the documents) or automatically, by using an unsupervised
algorithm to group similar documents in topic-like clusters. We have
utilized the latter approach, for its generality and extensibility,
and because there is no reason to believe that the manually assigned
topics are optimal for language modeling.

\subsection{Tree Generation}

In this study, we have investigated a range of hierarchical clustering
techniques, examining extensions of hierarchical agglomerative clustering,
\( k \)-means clustering and top-down EM-based clustering. The latter
underperformed on evaluations in \newcite{florian98} and is not reported
here.

A generic hierarchical agglomerative clustering algorithm proceeds
as follows: initially each document has its own cluster. Repeatedly,
the two closest clusters are merged and replaced by their union, until
there is only one top-level cluster. Pairwise document similarity
may be based on a range of functions, but to facilitate comparative
analysis we have utilized standard cosine similarity (\( d\left( D_{1},D_{2}\right) =\frac{\left\langle D_{1},D_{2}\right\rangle }{\left\Vert D_{1}\right\Vert _{2}\left\Vert D_{2}\right\Vert _{2}} \))
and IR-style term vectors (see \newcite{salton83}).

\begin{figure}[tb]
{\centering \begin{tabular}{ccc}
\resizebox*{0.12\textwidth}{0.15\textheight}{\rotatebox{-90}{\includegraphics{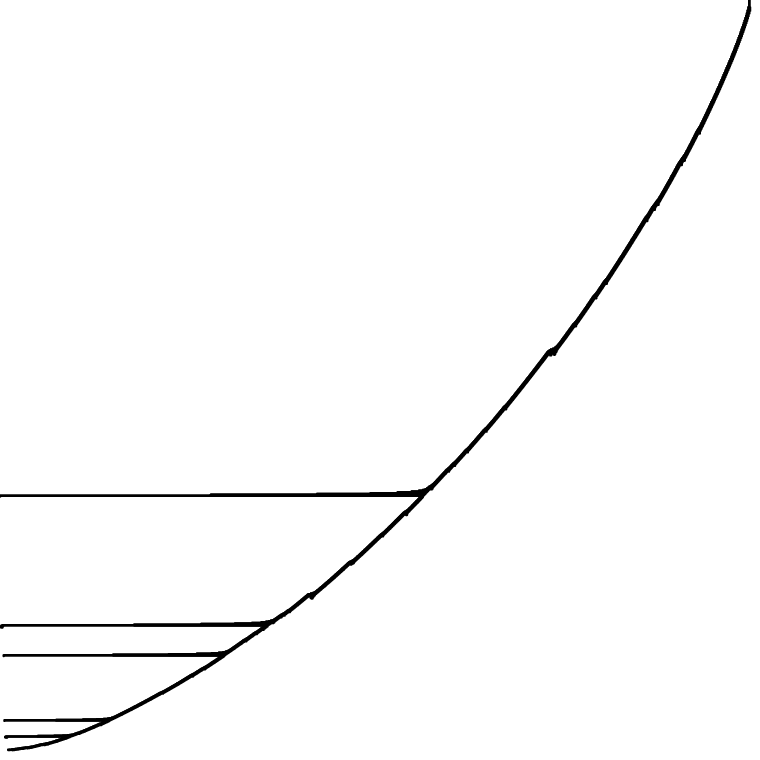}}} &
\resizebox*{0.12\textwidth}{0.15\textheight}{\rotatebox{-90}{\includegraphics{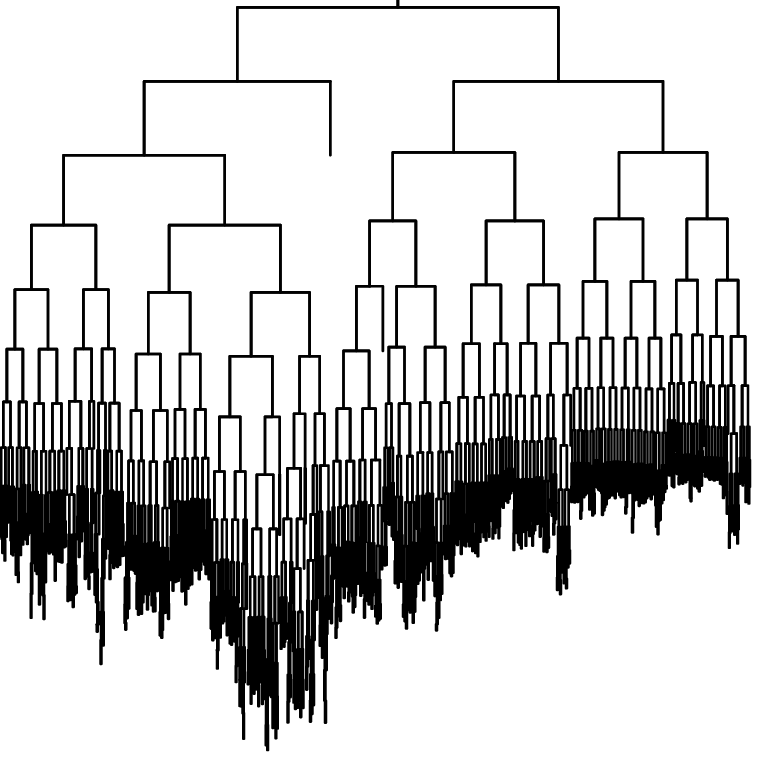}}} &
\resizebox*{0.12\textwidth}{0.15\textheight}{\rotatebox{-90}{\includegraphics{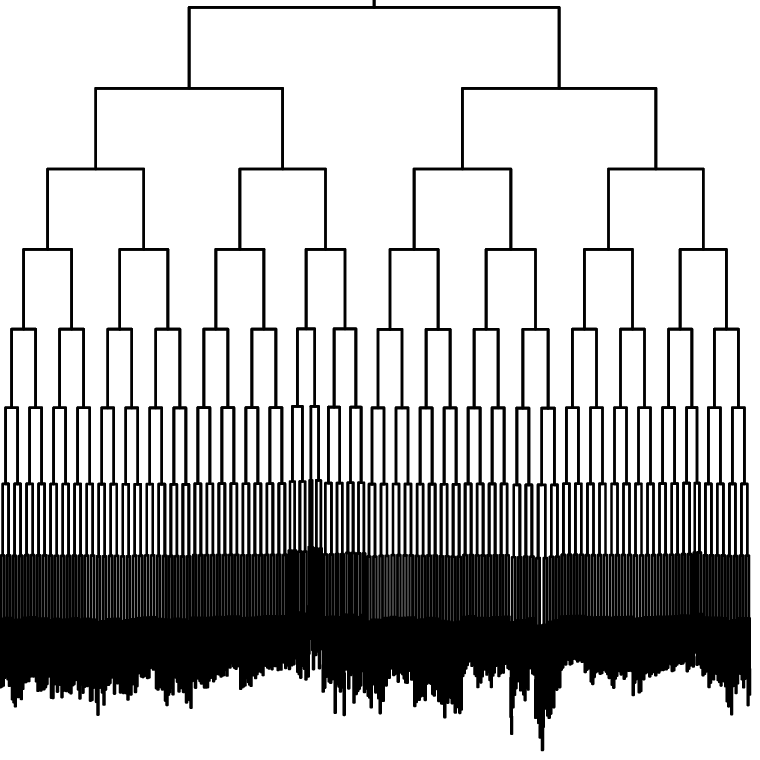}}} \\
\( \alpha =0 \)&
\( \alpha =0.3 \)&
\( \alpha =0.5 \)\\
\end{tabular}\par}

\caption{As \protect\( \alpha \protect \) increases, the trees become more
balanced, at the expense of forced clustering\label{table1}}
\end{figure}

This procedure outputs a tree in which documents on similar topics
(indicated by similar term content) tend to be clustered together.
The difference between average-linkage and maximum-linkage algorithms
manifests in the way the similarity between clusters is computed (see
\newcite{duda73}). A problem that appears when using hierarchical
clustering is that small centroids tend to cluster with bigger centroids
instead of other small centroids, often resulting in highly skewed
trees such as shown in Figure \ref{table1}, \( \alpha  \)=0. To
overcome the problem, we devised two alternative approaches for computing
the intercluster similarity:

\begin{itemize}
\item Our first solution minimizes the attraction of large clusters by introducing
a normalizing factor \( \alpha  \) to the inter-cluster distance
function: 
\end{itemize}
\vspace*{-0.2cm}{\small \begin{equation}
\label{21}
\vspace *{-2cm}d(C_{1},C_{2})=\frac{<\operatorname {c}(C_{1}),\operatorname {c}(C_{2})>}{N(C_{1})^{\alpha }\left\Vert \operatorname {c}(C_{1})\right\Vert N(C_{2})^{\alpha }\left\Vert \operatorname {c}(C_{2})\right\Vert }
\end{equation}
} 

\vspace*{-0.3cm}\begin{itemize}\item[] where \( N\left( C_{k}\right)  \)
is the number of vectors (documents) in cluster \( C_{k} \) and \( c\left( C_{i}\right)  \)
is the centroid of the \( i^{\textrm{th}} \) cluster. Increasing
\( \alpha  \) improves tree balance as shown in Figure 2, but as
\( \alpha  \) becomes large the forced balancing degrades cluster
quality. \footnotetext{ Section \ref{opimizesec} describes the choice
of optimum \( \alpha  \). }\end{itemize}

\begin{itemize}
\item A second approach we explored is to perform basic smoothing of term
vector weights, replacing all 0's with a small value \( \epsilon  \).
By decreasing initial vector orthogonality, this approach facilitates
attraction to small centroids, and leads to more balanced clusters
as shown in Figure 3. 
\end{itemize}

\begin{figure}[tb]
{\centering \begin{tabular}{cccc}
\resizebox*{0.1\textwidth}{0.2\textheight}{\rotatebox{-90}{\includegraphics{al.0.eps}}}  &
\resizebox*{0.1\textwidth}{0.2\textheight}{\rotatebox{-90}{\includegraphics{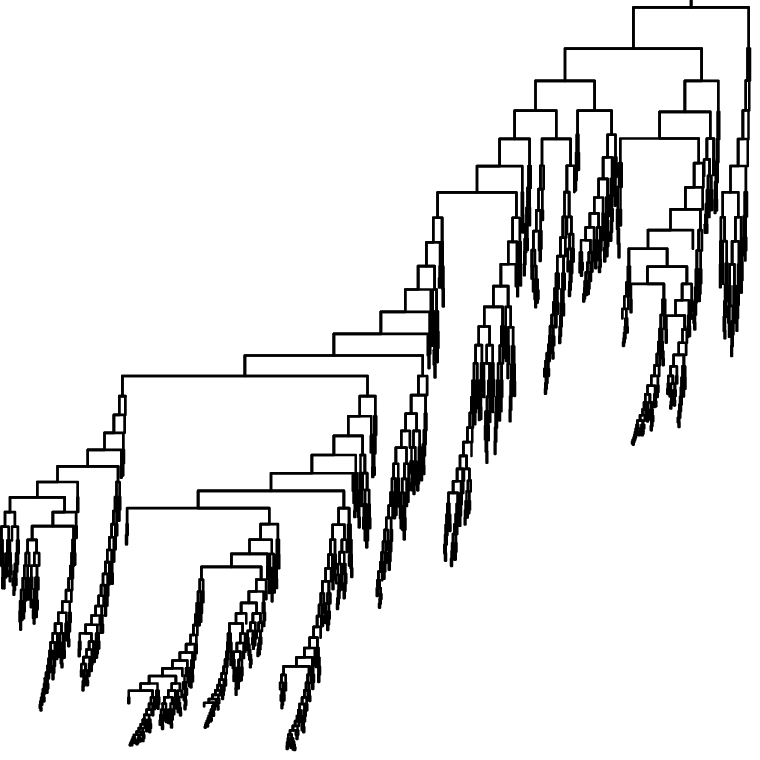}}}  &
\resizebox*{0.1\textwidth}{0.2\textheight}{\rotatebox{-90}{\includegraphics{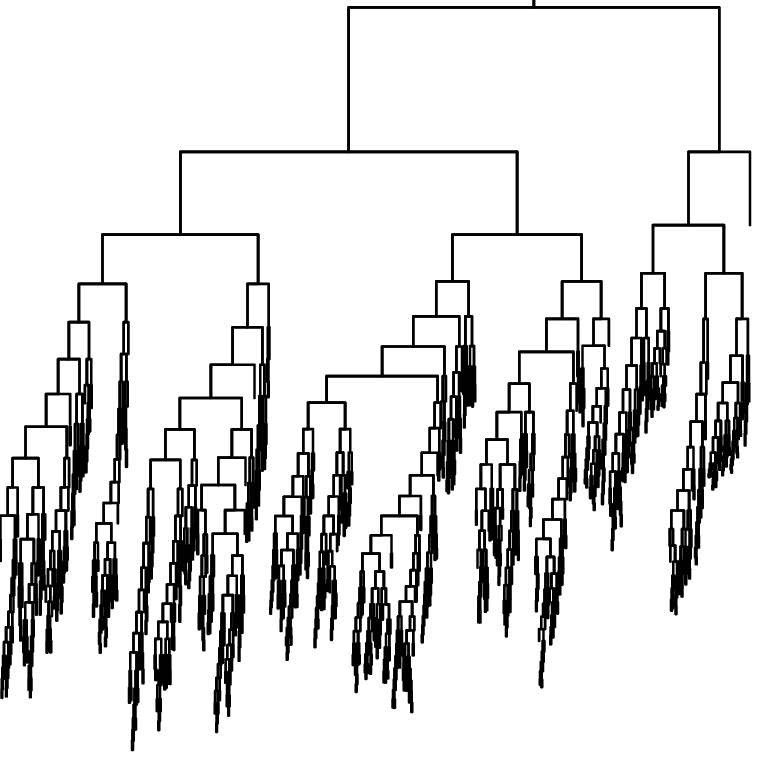}}}  &
\resizebox*{0.1\textwidth}{0.2\textheight}{\rotatebox{-90}{\includegraphics{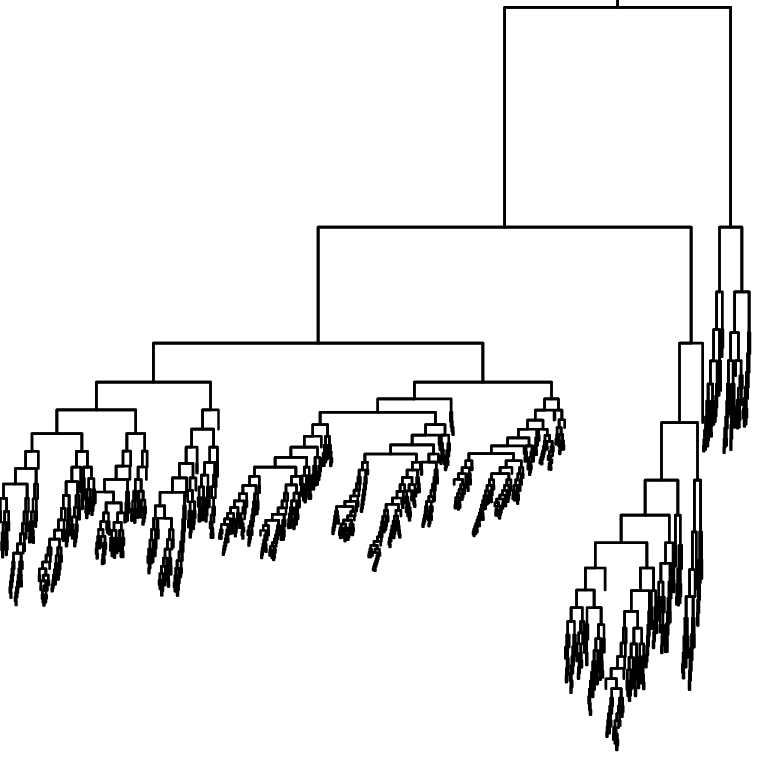}}} \\
 \( \epsilon =0 \)&
 \( \epsilon =0.15 \)&
 \( \epsilon =0.3 \)&
 \( \epsilon =0.7 \)\\
\end{tabular}\par}

\caption{Tree-balance is also sensitive to the smoothing parameter \protect\( \epsilon \protect \).\label{table2}}
\end{figure}

Instead of stopping the process when the desired number of clusters
is obtained, we generate the full tree for two reasons: (1) the full
hierarchical structure is exploited in our language models and (2)
once the tree structure is generated, the objective function we used
to partition the tree differs from that used when building the tree.
Since the clustering procedure turns out to be rather expensive for
large datasets (both in terms of time and memory), only 10000 documents
were used for generating the initial hierarchical structure.

\subsection{Optimizing the Hierarchical Structure\label{opimizesec}}

To be able to compute accurate language models, one has to have sufficient
data for the relative frequency estimates to be reliable. Usually,
even with enough data, a smoothing scheme is employed to insure that
{\footnotesize \( P\left( w_{i}|w_{1}^{i-1}\right) >0 \)} for any
given word sequence \( w_{1}^{i} \).

The trees obtained from the previous step have documents in the leaves,
therefore not enough word mass for proper probability estimation.
But, on the path from a leaf to the root, the internal nodes grow
in mass, ending with the root where the counts from the entire corpus
are stored. Since our intention is to use the full tree structure
to interpolate between the in-node language models, we proceeded to
identify a subset of internal nodes of the tree, which contain sufficient
data for language model estimation. The criteria of choosing the nodes
for collapsing involves a goodness function, such that the \emph{cut}\footnote{%
the collection of nodes that collapse 
} is a solution to a constrained optimization problem, given the constraint
that the resulting tree has exactly \( k \) leaves. Let this evaluation
function be \( g(n) \), where \( n \) is a node of the tree, and
suppose that we want to minimize it. Let \( g(n,k) \) be the minimum
cost of creating \( k \) leaves in the subtree of root \( n \).
When the evaluation function \( g\left( n\right)  \) satisfies the
locality condition that it depends solely on the values {\footnotesize \( g\left( n_{j},\cdot \right)  \),}
(where {\footnotesize \( \left( n_{j}\right) _{j=1..k} \)}are the
children of node \( n \)), \( g\left( root\right)  \) can be computed
efficiently using dynamic programming\footnote{%
\( h \) is an operator through which the values \( g\left( n_{1},j_{1}\right) ,\ldots ,g\left( n_{k},j_{k}\right)  \)
are combined, as \( \sum  \) or \( \prod  \)
} : 

{\small \begin{equation}
\label{344}
\begin{array}[t]{l}
g(n,1)=g(n)\\
g(n,k)=\min \limits_{\begin{array}{c}
j_{1},,j_{k}\geq 1\\
\sum _{k}j_{k}=k
\end{array}}h\left( g\left( n_{1},j_{1}\right) ,\ldots ,g\left( n_{k},j_{k}\right) \right) 
\end{array}
\end{equation}
}{\small \par}

\begin{figure*}[tb]
\centering

\resizebox*{0.4\textwidth}{0.2\textheight}{\includegraphics{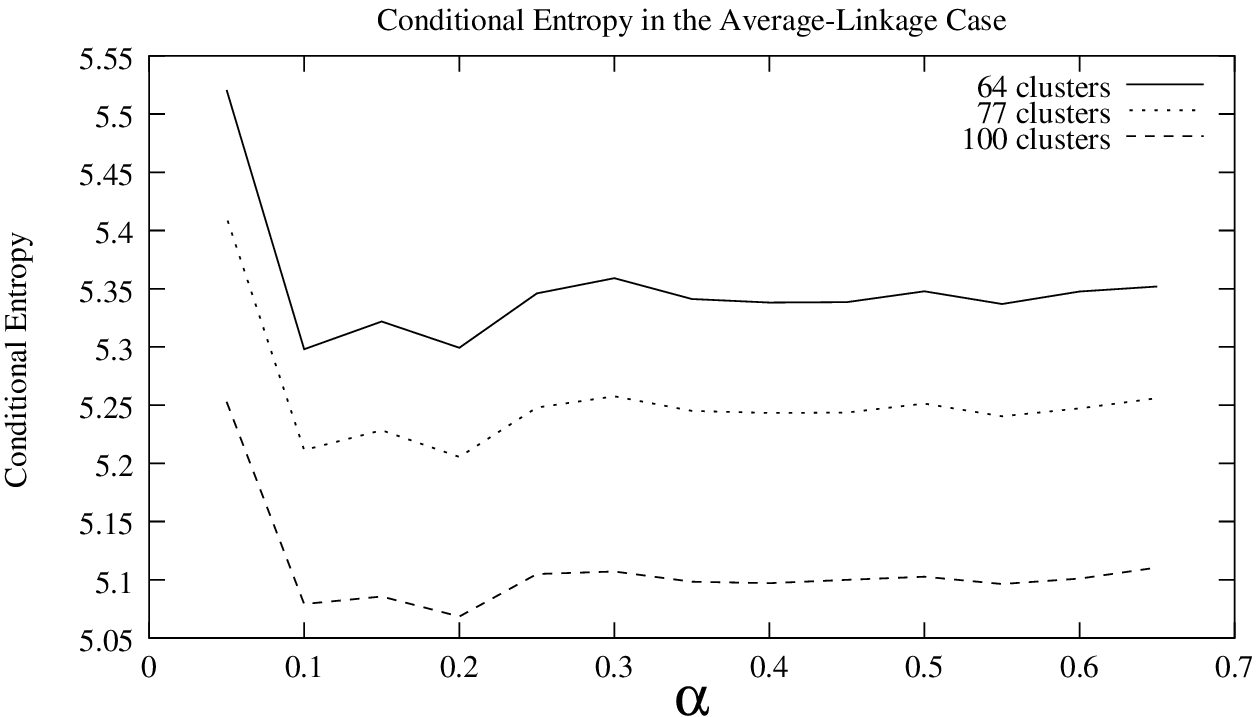}}  \resizebox*{0.4\textwidth}{0.2\textheight}{\includegraphics{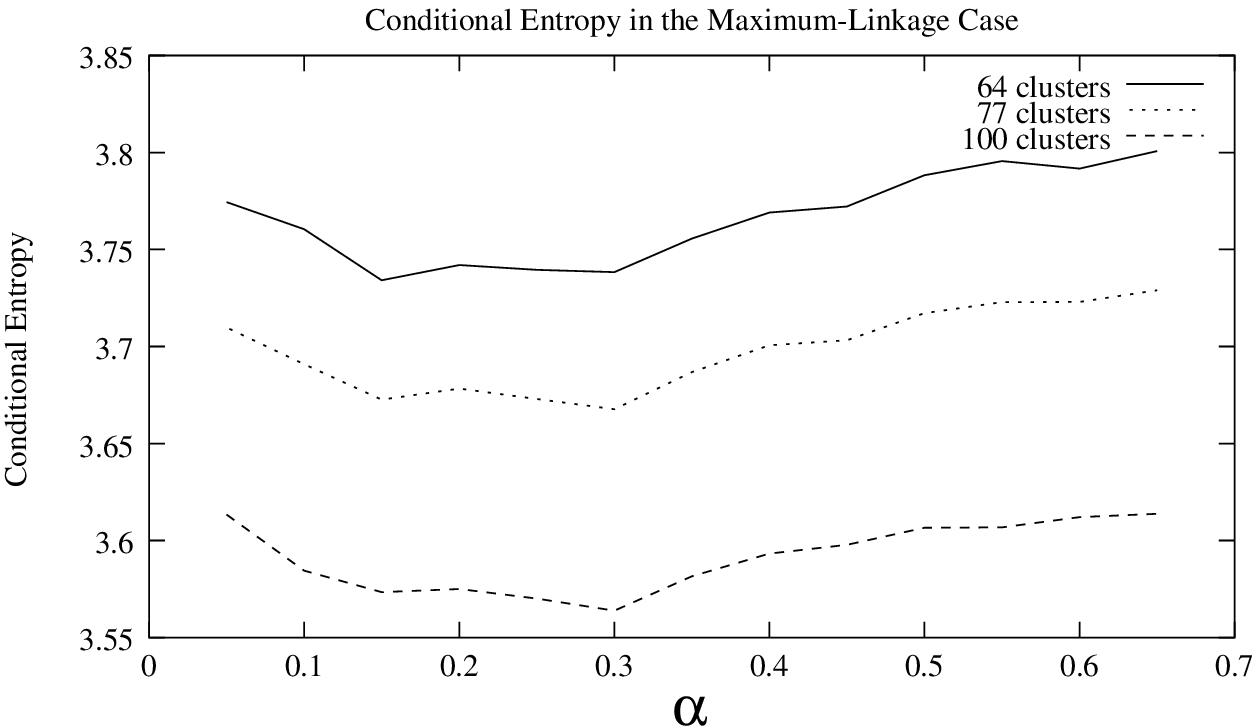}}

\caption{Conditional entropy for different \protect\( \alpha \protect \),
cluster sizes and linkage methods \label{fig4}}
\end{figure*}
Let us assume for a moment that we are interested in computing a unigram
topic-mixture language model. If the topic-conditional distributions
have high entropy (e.g. the histogram of \( P(w|topic) \) is fairly
uniform), topic-sensitive language model interpolation will not yield
any improvement, no matter how well the topic detection procedure
works. Therefore, we are interested in clustering documents in such
a way that the topic-conditional distribution \( P(w|topic) \) is
maximally skewed. With this in mind, we selected the evaluation function
to be the \emph{conditional entropy} of a set of words (possibly the
whole vocabulary) given the particular classification. The conditional
entropy of some set of words \( \mathcal{W} \) given a partition
\( \mathcal{C} \) is \begin{small}{\small \begin{equation}
\label{entropy}
\begin{array}{r}
H(\mathcal{W}|\mathcal{C})=\sum \limits_{i=1}^{n}P(C_{i})\sum \limits_{w\in \mathcal{W}\cap C_{i}}P(w|C_{i})\cdot \log (P(w|C_{i}))\\
=\frac{1}{T}\sum \limits_{i=1}^{n}\sum \limits_{w\in \mathcal{W}\cap C_{i}}c(w,C_{i})\cdot \log (P(w|C_{i}))
\end{array}\, \, \, \, \, \, 
\end{equation}
} \end{small}where \( c\left( w,C_{i}\right)  \) is the TF-IDF factor
of word \( w \) in class \( C_{i} \) and \( T \) is the size of
the corpus. Let us observe that the conditional entropy does satisfy
the locality condition mentioned earlier.

Given this objective function, we identified the optimal tree cut
using the dynamic-programming technique described above. We also optimized
different parameters (such as \( \alpha  \) and choice of linkage
method).

Figure \ref{fig4} illustrates that for a range of cluster sizes,
maximal linkage clustering with \( \alpha  \)=0.15-0.3 yields optimal
performance given the objective function in equation (\ref{21}).

The effect of varying \( \alpha  \) is also shown graphically in
Figure 5. Successful tree construction for language modeling purposes
will minimize the conditional entropy of \( P\left( \mathcal{W}|\mathcal{C}\right)  \).
This is most clearly illustrated for the word \textit{politics}, where
the tree generated with \( \alpha =0.3 \) maximally focuses documents
on this topic into a single cluster. The other words shown also exhibit
this desirable highly skewed distribution of \( P\left( \mathcal{W}|\mathcal{C}\right)  \)
in the cluster tree generated when \( \alpha =0.3 \).

Another investigated approach was k-means clustering (see \newcite{duda73})
as a robust and proven alternative to hierarchical clustering. Its
application, with both our automatically derived clusters and Mangu's
manually derived clusters (\newcite{Mangu97}) used as initial partitions,
actually yielded a small increase in conditional entropy and was not
pursued further.

\section{Language Model Construction and Evaluation\label{TheLanguageModel}}

Estimating the language model probabilities is a two-phase process.
First, the topic-sensitive language model probabilities {\footnotesize \( P\left( w_{i}|t,w_{i-m+1}^{i-1}\right)  \)}
are computed during the training phase. Then, at run-time, or in the
testing phase, topic is dynamically identified by computing the probabilities
{\footnotesize \( P\left( t|w_{1}^{i-1}\right)  \)} as in section
\ref{dyntop} and the final language model probabilities are computed
using Equation (\ref{12}). The tree used in the following experiments
was generated using average-linkage agglomerative clustering, using
parameters that optimize the objective function in Section \ref{optimizesec}.

\subsection{Language Model Construction}

\begin{figure*}[tb]
\begin{itemize}
\item {\footnotesize Case 1: \( \operatorname {f}_{\textrm{node}}\left( w_{1}\right) \neq 0 \) }{\footnotesize \par}

{\footnotesize {\centering\( \hat{P}_{\textrm{node}}\left( w_{2}|w_{1}\right) =\left\{ \begin{array}{llll}
P_{\textrm{root}}\left( w_{2}|w_{1}\right)  & \textrm{if }w_{2}\in \mathcal{F}\left( w_{1}\right)  &  & \\
\lambda _{1}\operatorname {f}_{\textrm{node}}\left( w_{2}|w_{1}\right) \cdot \gamma _{\textrm{node}}\left( w_{1}\right) +\lambda _{2}\hat{P}_{\textrm{node}}\left( w_{2}\right)  &  &  & \\
\, \, \, \, \, \, \, \, \, \, \, \, \, \, \, \, \, \, +\left( 1-\lambda _{1}-\lambda _{2}\right) \hat{P}_{\textrm{parent}(\textrm{node})}\left( w_{2}|w_{1}\right)  & \textrm{if }w_{2}\in \mathcal{R}\left( w_{1}\right)  &  & \\
\alpha _{\textrm{node}}\left( w_{1}\right) \hat{P}_{\textrm{node}}\left( w_{2}\right)  & \textrm{if }w_{2}\in \mathcal{U}\left( w1\right)  & 
\end{array}\right.  \) }}{\footnotesize \par}

{\footnotesize where }{\footnotesize \par}

{\footnotesize \( \begin{array}{cc}
\gamma _{\textrm{node}}\left( w_{1}\right) =\frac{1-\sum\limits _{w_{2}\in \mathcal{F}\left( w_{1}\right) }\operatorname {f}_{\textrm{node}}\left( w_{2}|w_{1}\right) }{\left( 1+\beta \right) \sum\limits _{w_{2}\in \mathcal{R}\left( w_{1}\right) }\operatorname {f}_{\textrm{node}}\left( w_{2}|w_{1}\right) },\, \alpha _{\textrm{node}}\left( w_{1}\right) =\frac{\beta \left( 1-\sum\limits _{w_{2}\in \mathcal{F}\left( w_{1}\right) }\operatorname {f}_{\textrm{node}}\left( w_{2}|w_{1}\right) \right) }{\left( 1+\beta \right) \left( 1-\sum\limits _{w_{2}\in \mathcal{F}\left( w_{1}\right) \cup \mathcal{R}\left( w_{1}\right) }\hat{P}_{\textrm{node}}\left( w_{2}\right) \right) } & \\

\end{array} \)}{\footnotesize \par}

\item {\footnotesize Case 2: \( \operatorname {f}_{\textrm{node}}\left( w_{1}\right) =0 \)
\[
\hat{P}_{\textrm{node}}\left( w_{2}|w_{1}\right) =\left\{ \begin{array}{llll}
P_{\textrm{root}}\left( w_{2}|w_{1}\right)  & \textrm{if }w_{2}\in \mathcal{F}\left( w_{1}\right)  &  & \\
\lambda _{2}\hat{P}_{\textrm{node}}\left( w_{2}\right) \cdot \gamma _{\textrm{node}}\left( w_{1}\right)  &  &  & \\
\, \, \, \, \, +\left( 1-\lambda _{3}\right) \hat{P}_{\textrm{parent}(\textrm{node})}\left( w_{2}|w_{1}\right)  & \textrm{if }w_{2}\in \mathcal{R}\left( w_{1}\right)  &  & \\
\alpha _{\textrm{node}}\left( w_{1}\right) \hat{P}_{\textrm{node}}\left( w_{2}\right)  & \textrm{if }w_{2}\in \mathcal{U}\left( w1\right)  & 
\end{array}\right. \]
 where \( \gamma _{\textrm{node}}\left( w_{1}\right)  \) and \( \alpha _{\textrm{node}}\left( w_{1}\right)  \)
are computed in a similar fashion such that the probabilities do sum
to \( 1 \). }{\footnotesize \par}
\end{itemize}

\caption{Basic Bigram Language Model Specifications\label{bigram-formulae}}
\end{figure*}
The topic-specific language model probabilities are computed in a
four phase process:

\begin{enumerate}
\item Each document is assigned to one leaf in the tree, based on the similarity
to the leaves' centroids (using the cosine similarity). The document
counts are added to the selected leaf's count. 
\item The leaf counts are propagated up the tree such that, in the end,
the counts of every internal node are equal to the sum of its children's
counts. At this stage, each node of the tree has an attached language
model - the relative frequencies. 
\item In the root of the tree, a discounted Good-Turing language model is
computed (see \newcite{katz87}, \newcite{chen98}). 
\item \( m \)-gram smooth language models are computed for each node \( n \)
different than the root by three-way interpolating between the \( m \)-gram
language model in the parent \( parent(n) \), the \( \left( m-1\right)  \)-gram
smooth language model in node \( n \) and the \( m \)-gram relative
frequency estimate in node \( n \):  {\small \begin{equation}
\label{42}
\begin{array}{l}
\hat{P}_{n}\left( w_{m}|w_{1}^{m-1}\right) =\\
\begin{array}{r}
\lambda _{n}^{1}\left( w_{1}^{m-1}\right) \hat{P}_{\textrm{parent}(n)}\left( w_{m}|w_{1}^{m-1}\right) \\
+\lambda _{n}^{2}\left( w_{1}^{m-1}\right) \hat{P}_{n}\left( w_{m}|w_{2}^{m-1}\right) \\
+\lambda _{n}^{3}\left( w_{1}^{m-1}\right) f_{n}\left( w_{m}|w_{1}^{m-1}\right) 
\end{array}
\end{array}
\end{equation}
}with {\small \( \lambda _{n}^{1}\left( w_{1}^{m-1}\right) +\lambda _{n}^{2}\left( w_{1}^{m-1}\right) +\lambda _{n}^{3}\left( w_{1}^{m-1}\right) =1 \)}
for each node \( n \) in the tree. Based on how {\small \( \lambda _{n}^{k}\left( w_{1}^{m-1}\right)  \)}
depend on the particular node \( n \) and the word history {\small \( w_{1}^{m-1} \)},
various models can be obtained. We investigated two approaches: a
bigram model in which the \( \lambda  \)'s are fixed over the tree,
and a more general trigram model in which \( \lambda 's \) adapt
using an EM reestimation procedure.
\end{enumerate}

\subsubsection{Bigram Language Model}

Not all words are topic sensitive. \newcite{Mangu97} observed that
closed-class function words (FW), such as \textit{the}, \textit{of},
and \textit{with}, have minimal probability variation across different
topic parameterizations, while most open-class content words (CW)
exhibit substantial topic variation. This leads us to divide the possible
word pairs in two classes (topic-sensitive and not) and compute the
\( \lambda  \)'s in Equation (\ref{42}) in such a way that the probabilities
in the former set are constant in all the models. To formalize this: 

\begin{itemize}
\item \( \mathcal{F}\left( w_{1}\right) =\left\{ w_{2}\in \mathcal{V}|\left( w_{1},w_{2}\right) \textrm{ is fixed}\right\}  \)-the
{}``fixed{}'' space; 
\item \( \mathcal{R}\left( w_{1}\right) =\left\{ w_{2}\in \mathcal{V}|\left( w_{1},w_{2}\right) \textrm{ is free}/\textrm{variable}\right\}  \)-
the {}``free{}'' space; 
\item \( \mathcal{U}\left( w_{1}\right) =\left\{ w_{2}\in \mathcal{V}|\left( w_{1},w_{2}\right) \textrm{ was never seen}\right\}  \)-
the {}``unknown{}'' space. 
\end{itemize}
The imposed restriction is, then: for every word \( w_{1} \)and any
word \( w_{2}\in \mathcal{F}\left( w_{1}\right)  \) \( P_{n}\left( w_{2}|w_{1}\right) =P_{root}\left( w_{2}|w_{1}\right)  \)
in any node \( n. \)

The distribution of bigrams in the training data is as follows, with
roughly 30\% bigram probabilities allowed to vary in the topic-sensitive
models:

This approach raises one interesting issue: the language model in
the root assigns some probability mass to the unseen events, equal
to the singletons' mass (see \newcite{good53},\newcite{katz87}). In
our case, based on the assumptions made in the Good-Turing formulation,
we considered that the ratio of the probability mass that goes to
the unseen events and the one that goes to seen, free events should
be fixed over the nodes of the tree. Let \( \beta  \) be this ratio.
Then the language model probabilities are computed as in Figure \ref{bigram-formulae}. 
\begin{table}[tb]
\begin{tabular}{|c|c|c|c|c|}
\hline 
{\tiny Model }&
 {\tiny Bigram-type }&
 {\tiny Example }&
{\tiny Freq. }&
\\
\hline
{\tiny fixed}&
 {\tiny \( p(FW|FW) \)}&
 {\tiny \( p(the|in) \)}&
 {\tiny 45.3\%}&
 {\tiny least topic sensitive}\\
\hline
{\tiny fixed}&
 {\tiny \( p(FW|CW) \)}&
 {\tiny \( p(of|scenario) \)}&
 {\tiny 24.8\%}&
 {\tiny \( \downarrow  \)}\\
\hline
{\tiny free}&
 {\tiny \( p(CW|CW) \)}&
 {\tiny \( p(air|cold) \)}&
 {\tiny 5.3\%}&
 {\tiny \( \downarrow  \)}\\
\hline
{\tiny free}&
 {\tiny \( p(CW|FW) \)}&
 {\tiny \( p(air|the) \)}&
 {\tiny 24.5\%}&
 {\tiny most topic sensitive}\\
\hline
\end{tabular}
\end{table}

\subsubsection{Ngram Language Model Smoothing}

In general, \( n \) gram language model probabilities can be computed
as in formula (\ref{42}), where {\footnotesize \( \left( \lambda _{n}^{k}\left( w_{1}^{m-1}\right) \right) _{k=1\ldots 3} \)}
are adapted both for the particular node \( n \) and history {\footnotesize \( w_{1}^{m-1} \)}.
The proposed dependency on the history is realized through the history
count {\footnotesize \( C\left( w_{1}^{m-1}\right)  \)} and the relevance
of the history {\footnotesize \( w_{1}^{m-1} \)} to the topic in
the nodes \( n \) and \( parent\left( \textrm{n}\right)  \). The
intuition is that if a history is as relevant in the current node
as in the parent, then the estimates in the parent should be given
more importance, since they are better estimated. On the other hand,
if the history is much more relevant in the current node, then the
estimates in the node should be trusted more. The mean adapted \( \lambda  \)
for a given height \( h \) is the tree is shown in 
\begin{figure}[tb]
{\centering \resizebox*{0.8\columnwidth}{!}{\includegraphics{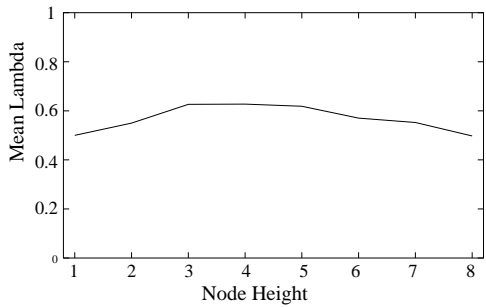}} \par}

\caption{Mean of the estimated \protect\( \lambda \protect \)s at node height
\protect\( h\protect \), in the unigram case\label{lambda}}
\end{figure}
Figure \ref{lambda}. This is consistent with the observation that
splits in the middle of the tree tend to be most informative, while
those closer to the leaves suffer from data fragmentation, and hence
give relatively more weight to their parent. As before, since not
all the \( m \)-grams are expected to be topic-sensitive, we use
a method to insure that those \( m \) grams are kept {}``fixed{}''
to minimize noise and modeling effort. In this case, though, 2 language
models with different support are used: one that supports the topic
insensitive \( m \)-grams and that is computed only once (it's a
normalization of the topic-insensitive part of the overall model),
and one that supports the rest of the mass and which is computed by
interpolation using formula (\ref{42}). Finally, the final language
model in each node is computed as a mixture of the two. 
\begin{figure*}[tb]
{\centering \resizebox*{0.96\textwidth}{0.35\textheight}{\includegraphics{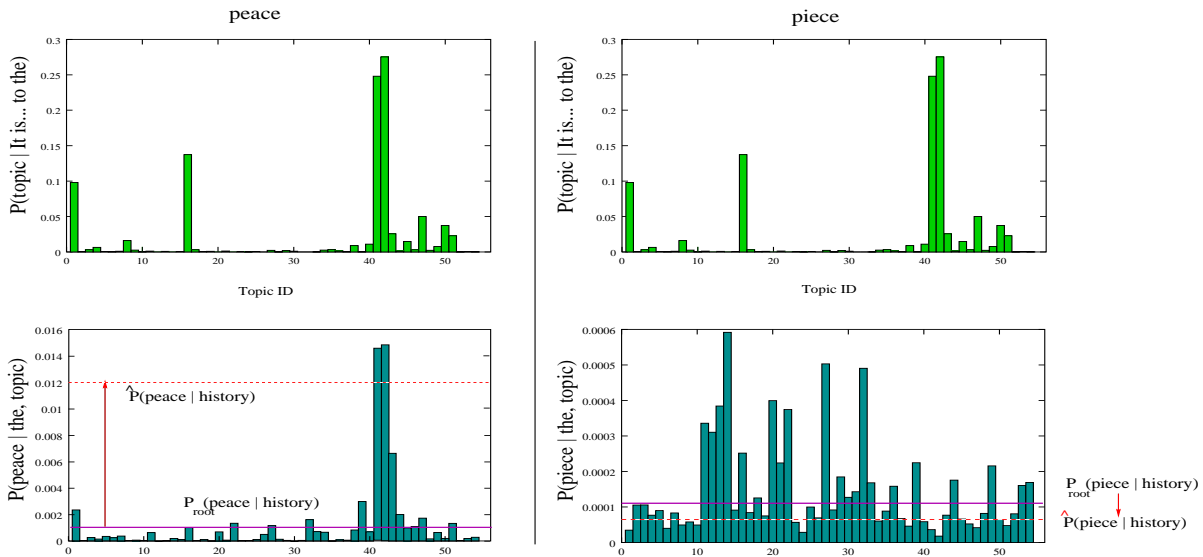}} \par}

\caption{Topic sensitive probability estimation for \textbf{peace\label{policeslide}}
and \textbf{piece} in context}
\end{figure*}

\subsection{Dynamic Topic Adaptation \label{dyntop} }

Consider the example of predicting the word following the Broadcast
News fragment: {}``It is at least on the Serb side a real drawback
to the \fbox{\ ? }{}''. Our topic detection model, as further detailed
later in this section, assigns a topic distribution to this left context
(including the full previous discourse), illustrated in the upper
portion of Figure \ref{policeslide}. The model identifies that this
particular context has greatest affinity with the empirically generated
topic clusters \#41 and \#42 (which appear to have one of their foci
on international events).

\begin{table*}[tb]
\centering 

\begin{tabular}{|c|c|c|c|c|c|c|c|}
\hline 
\multicolumn{6}{|c|}{\textbf{Language} }&
 \textbf{Perplexity} \textbf{on}&
 \textbf{Perplexity} \textbf{on}\\
\multicolumn{6}{|c|}{\textbf{Model }}&
 \textbf{the entire}&
 \textbf{the target}\\
\multicolumn{6}{|c|}{}&
 \textbf{vocabulary}&
 \textbf{vocabulary}\\
\hline
\multicolumn{6}{|c|}{\textbf{Standard Bigram Model}}&
 \textbf{215}&
\multicolumn{1}{|c|}{\textbf{584}}\\
\hline
&
\multicolumn{1}{c|}{ \textbf{History size}}&
\multicolumn{1}{c|}{ \textbf{Scaled}}&
\multicolumn{1}{c|}{ \( g\left( x\right)  \)}&
\multicolumn{1}{c|}{ \( f\left( x\right)  \)}&
\multicolumn{1}{c|}{ k-NN}&
\multicolumn{2}{|c|}{ }\\
\cline{2-2} \cline{3-3} \cline{4-4} \cline{5-5} \cline{6-6} \cline{7-8} 
\multicolumn{1}{|c|}{}&
 100&
 yes&
 \( x \)&
 \( x^{2} \)&
 -&
 206&
\multicolumn{1}{|c|}{460}\\
\cline{2-2} \cline{3-3} \cline{4-4} \cline{5-5} \cline{6-6} \cline{7-7} \cline{8-8} 
&
\multicolumn{1}{|c|}{1000}&
\multicolumn{1}{|c|}{yes}&
\multicolumn{1}{|c|}{\( x \)}&
\multicolumn{1}{c|}{ \( x^{2} \)}&
\multicolumn{1}{|c|}{-}&
\multicolumn{1}{|c|}{195}&
\multicolumn{1}{|c|}{405}\\
\cline{2-2} \cline{3-3} \cline{4-4} \cline{5-5} \cline{6-6} \cline{7-7} \cline{8-8} 
&
\multicolumn{1}{|c|}{5000}&
\multicolumn{1}{|c|}{yes\( ^{*} \)}&
\multicolumn{1}{|c|}{\( x \)\( ^{*} \)}&
\multicolumn{1}{|c|}{\( x^{2} \)\( ^{*} \)}&
\multicolumn{1}{|c|}{-\( ^{*} \)}&
\multicolumn{1}{|r|}{\textbf{192} (-10\%)}&
\multicolumn{1}{|r|}{\textbf{389}(-33\%)}\\
\cline{2-2} \cline{3-3} \cline{4-4} \cline{5-5} \cline{6-6} \cline{7-7} \cline{8-8} 
&
\multicolumn{1}{|c|}{5000}&
\multicolumn{1}{|c|}{yes}&
\multicolumn{1}{|c|}{\( 1 \)}&
\multicolumn{1}{|c|}{\( x \)}&
\multicolumn{1}{|c|}{-}&
\multicolumn{1}{|c|}{202}&
\multicolumn{1}{|c|}{444}\\
\cline{2-2} \cline{3-3} \cline{4-4} \cline{5-5} \cline{6-6} \cline{7-7} \cline{8-8} 
&
\multicolumn{1}{|c|}{5000}&
\multicolumn{1}{|c|}{no}&
\multicolumn{1}{|c|}{\( x \)}&
\multicolumn{1}{|c|}{\( x^{2} \)}&
\multicolumn{1}{|c|}{-}&
\multicolumn{1}{|c|}{193}&
\multicolumn{1}{|c|}{394}\\
\cline{2-2} \cline{3-3} \cline{4-4} \cline{5-5} \cline{6-6} \cline{7-7} \cline{8-8} 
&
\multicolumn{1}{|c|}{5000}&
\multicolumn{1}{|c|}{yes}&
\multicolumn{1}{|c|}{\( x \)}&
\multicolumn{1}{|c|}{\( x^{2} \)}&
\multicolumn{1}{|c|}{15-NN}&
\multicolumn{1}{|c|}{192}&
\multicolumn{1}{|c|}{390}\\
\cline{2-2} \cline{3-3} \cline{4-4} \cline{5-5} \cline{6-6} \cline{7-7} \cline{8-8} 
  \hspace*{2mm}\begin{rotate}{90}\hspace*{5mm}\textbf{Topic LMs} \end{rotate}&
 5000&
 yes&
 \( e^{x} \)&
 \( xe^{x} \)&
 -&
 196&
 411 \\
\hline
\end{tabular}

\caption{Perplexity results for topic sensitive bigram language model, different
history lengths\label{bigramperp}}
\end{table*}
The lower portion of Figure \ref{policeslide} illustrates the topic-conditional
bigram probabilities \( P(w|\textrm{the},topic) \) for two candidate
hypotheses for \textit{w}: \emph{peace} (the actually observed word
in this case) and \emph{piece} (an incorrect competing hypothesis).
In the former case, \( P(peace|\textrm{the},topic) \) is clearly
highly elevated in the most probable topics for this context (\#41,\#42),
and thus the application of our core model combination (Equation \ref{12})
yields a posterior joint product {\footnotesize \( P\left( w_{i}|w_{1}^{i-1}\right) =\sum _{t=1}^{K}P\left( t|w_{1}^{i-1}\right) \cdot P_{t}\left( w_{i}|w_{i-m+1}^{i-1}\right)  \)}
that is 12-times more likely than the overall bigram probability,
\( P(\textrm{air}|\textrm{the})=0.001 \). In contrast, the obvious
accustically motivated alternative \emph{piece,} has greatest probability
in a far different and much more diffuse distribution of topics, yielding
a joint model probability for this particular context that is \( 40\% \)
lower than its baseline bigram probability. This context-sensitive
adaptation illustrates the efficacy of dynamic topic adaptation in
increasing the model probability of the truth. 

Clearly the process of computing the topic detector {\small \( P\left( t|w_{1}^{i-1}\right)  \)}
is crucial. We have investigated several mechanisms for estimating
this probability, the most promising is a class of normalized transformations
of traditional cosine similarity between the document history vector
{\small \( w_{1}^{i-1} \)} and the topic centroids: {\small \begin{equation}
\label{probestim}
P\left( t|w_{1}^{i-1}\right) =\frac{f\left( \operatorname {Cosine-Sim}\left( t,w_{1}^{i-1}\right) \right) }{\sum\limits _{t^{\prime }}f\left( \operatorname {Cosine-Sim}\left( t^{\prime },w_{1}^{i-1}\right) \right) }
\end{equation}
} One obvious choice for the function \( f \) would be the identity.
However, considering a linear contribution of similarities poses a
problem: because topic detection is more accurate when the history
is long, even unrelated topics will have a non-trivial contribution
to the final probability\footnote{%
Due to unimportant word co-occurrences
}, resulting in poorer estimates.

One class of transformations we investigated, that directly address
the previous problem, adjusts the similarities such that closer topics
weigh more and more distant ones weigh less. Therefore, \( f \) is
chosen such that 

{\small \begin{equation}
\label{122}
\begin{array}{c}
\frac{f\left( x_{1}\right) }{f\left( x_{2}\right) }\leq \frac{x_{1}}{x_{2}}\textrm{ for }x_{1}\leq x_{2}\Leftrightarrow \\
\frac{f\left( x_{1}\right) }{x_{1}}\leq \frac{f\left( x_{2}\right) }{x_{2}}\textrm{ for }x_{1}\leq x_{2}
\end{array}
\end{equation}
t}hat is, \( \frac{f\left( x\right) }{x} \) should be a monotonically
increasing function on the interval \( \left[ 0,1\right]  \), or,
equivalently \( f\left( x\right) =x\cdot g\left( x\right)  \), \( g \)
being an increasing function on \( \left[ 0,1\right]  \). Choices
for \( g(x) \) include \( x \), \( x^{\gamma }(\gamma >0) \), \( log\left( x\right)  \),
\( e^{x} \).

Another way of solving this problem is through the scaling operator
\( f'\left( x_{i}\right) =\frac{x_{i}-\min x_{i}}{\max x_{i}-\min x_{i}} \).
By applying this operator, minimum values (corresponding to low-relevancy
topics) do not receive any mass at all, and the mass is divided between
the more relevant topics. For example, a combination of scaling and
\( g(x)=x^{\gamma } \) yields:

\begin{equation}
\label{44}
\begin{array}{l}
P\left( t_{j}|w_{1}^{i-1}\right) =\\
\frac{\left( \frac{Sim\left( w_{1}^{i-1},t_{j}\right) -\min _{k}Sim\left( w_{1}^{i-1},t_{k}\right) }{\max _{k}Sim\left( w_{1}^{i-1},t_{k}\right) -\min _{k}Sim\left( w_{1}^{i-1},t_{k}\right) }\right) ^{\gamma }}{\sum\limits _{l}\left( \frac{Sim\left( w_{1}^{i-1},t_{l}\right) -\min _{k}Sim\left( w_{1}^{i-1},t_{k}\right) }{\max _{k}Sim\left( w_{1}^{i-1},t_{k}\right) -\min _{k}Sim\left( w_{1}^{i-1},t_{k}\right) }\right) ^{\gamma }}
\end{array}
\end{equation}

A third class of transformations we investigated considers only the
closest \( k \) topics in formula (\ref{probestim}) and ignores
the more distant topics.

\subsection{Language Model Evaluation}

Table \ref{bigramperp} briefly summarizes a larger table of performance
measured on the bigram implementation of this adaptive topic-based
LM. For the default parameters (indicated by \( ^{*} \)), a statistically
significant overall perplexity decrease of 10.5\% was observed relative
to a standard bigram model measured on the same 1000 test documents.
Systematically modifying these parameters, we note that performance
is decreased by using shorter discourse contexts (as histories never
cross discourse boundaries, 5000-word histories essentially correspond
to the full prior discourse). Keeping other parameters constant, \( g(x)=x \)
outperforms other candidate transformations \( g(x)=1 \) and \( g(x)=e^{x} \).
Absence of k-nn and use of scaling both yield minor performance improvements.

It is important to note that for 66\% of the vocabulary the topic-based
LM is identical to the core bigram model. On the 34\% of the data
that falls in the model's target vocabulary, however, perplexity reduction
is a much more substantial 33.5\% improvement. The ability to isolate
a well-defined target subtask and perform very well on it makes this
work especially promising for use in model combination.

\section{Conclusion }

In this paper we described a novel method of generating and applying
hierarchical, dynamic topic-based language models. Specifically, we
have proposed and evaluated hierarchical cluster generation procedures
that yield specially balanced and pruned trees directly optimized
for language modeling purposes. We also present a novel hierarchical
interpolation algorithm for generating a language model from these
trees, specializing in the hierarchical topic-conditional probability
estimation for a target topic-sensitive vocabulary (34\% of the entire
vocabulary). We also propose and evaluate a range of dynamic topic
detection procedures based on several transformations of content-vector
similarity measures. These dynamic estimations of \( P(topic_{i}|history) \)
are combined with the hierarchical estimation of \( P(word_{j}|topic_{i},history) \)
in a product across topics, yielding a final probability estimate
of \( P(word_{j}|history) \) that effectively captures long-distance
lexical dependencies via these intermediate topic models. Statistically
significant reductions in perplexity are obtained relative to a baseline
model, both on the entire text (\( 10.5\% \)) and on the target vocabulary
(\( 33.5\% \)). This large improvement on a readily isolatable subset
of the data bodes well for further model combination.

\section*{Acknowledgements}

The research reported here was sponsored by National Science Foundation
Grant IRI-9618874. The authors would like to thank Eric Brill, Eugene
Charniak, Ciprian Chelba, Fred Jelinek, Sanjeev Khudanpur, Lidia Mangu
and Jun Wu for suggestions and feedback during the progress of this
work, and Andreas Stolcke for use of his hierarchical clustering tools
as a basis for some of the clustering software developed here.

\bibliographystyle{acl}
\bibliography{citations}

\end{document}